\documentclass[lettersize,journal]{IEEEtran}
\usepackage{textcomp}
\usepackage{verbatim}
\usepackage{graphicx}
\usepackage{enumitem}

\usepackage{epsf}
\usepackage{epsfig}
\usepackage{epstopdf}
\usepackage{ifpdf}
\usepackage{cite}
\usepackage[utf8]{luainputenc}
\usepackage{amssymb}
\usepackage{amsthm}
\usepackage{lipsum}

\usepackage{dsfont}

\usepackage{pifont}

\ifCLASSINFOpdf
\else
\fi

\usepackage[cmex10]{amsmath}
\usepackage{amsmath}

\usepackage{array}
\usepackage{mdwmath}
\usepackage{mdwtab}
\usepackage{eqparbox}
\usepackage{url}

\usepackage{breakurl}
\usepackage[breaklinks]{hyperref}
\usepackage{caption}
\usepackage{subcaption}
\usepackage{amsfonts}
\usepackage{bbm}
\usepackage{breakurl}
\usepackage{doi}
\usepackage{float}
\usepackage{epsf}
\usepackage{epsfig}
\usepackage{epstopdf}
\usepackage{algorithmic}
\usepackage[linesnumbered, ruled, vlined]{algorithm2e}
\SetKw{KwOr}{or}
\SetKw{KwAnd}{and}
\SetKw{Cnt}{continue}
\usepackage{xcolor}

\begin{document}
\title{ORAN-GUIDE: RAG-Driven Prompt Learning for LLM-Augmented Reinforcement Learning in O-RAN Network Slicing \thanks{This material is based upon work supported by the National Science Foundation under Grant Numbers  CNS-2202972, CNS- 2318726, and CNS-2232048.} 
}

\author{Fatemeh Lotfi,~\IEEEmembership{member,~IEEE}, 
Hossein Rajoli,~\IEEEmembership{member,~IEEE}, Fatemeh Afghah~\IEEEmembership{Senior member,~IEEE}

\thanks{F. Lotfi, H. Rajoli and F. Afghah are with the Department of Electrical \& Computer Engineering, Clemson University e-mail: flotfi@clemson.edu, hrajoli@clemson.edu, fafghah@clemson.edu.}
}


\markboth{Journal of \LaTeX\ Class Files,~Vol.~14, No.~8, August~2021}%
{Shell \MakeLowercase{\textit{et al.}}: A Sample Article Using IEEEtran.cls for IEEE Journals}


\maketitle

\begin{abstract}

Advanced wireless networks must support highly dynamic and heterogeneous service demands. Open Radio Access Network (O-RAN) architecture enables this flexibility by adopting modular, disaggregated components, such as the RAN Intelligent Controller (RIC), Centralized Unit (CU), and Distributed Unit (DU), that can support intelligent control via machine learning (ML). While deep reinforcement learning (DRL) is a powerful tool for managing dynamic resource allocation and slicing, it often struggles to process raw, unstructured input like RF features, QoS metrics, and traffic trends. These limitations hinder policy generalization and decision efficiency in partially observable and evolving environments. 
To address this, we propose \textit{ORAN-GUIDE}, a dual-LLM framework that enhances multi-agent RL (MARL) with task-relevant, semantically enriched state representations. The architecture employs a domain-specific language model, ORANSight, pretrained on O-RAN control and configuration data, to generate structured, context-aware prompts. These prompts are fused with learnable tokens and passed to a frozen GPT-based encoder that outputs high-level semantic representations for DRL agents. This design adopts a retrieval-augmented generation (RAG) style pipeline tailored for technical decision-making in wireless systems. Experimental results show that ORAN-GUIDE improves sample efficiency, policy convergence, and performance generalization over standard MARL and single-LLM baselines. 



\end{abstract}

\begin{IEEEkeywords}
Open RAN, Network Slicing, Deep Reinforcement Learning, Retrieval-augmented generation (RAG), knowledge distillation.
\end{IEEEkeywords}

\IEEEpeerreviewmaketitle

\section{Introduction}
\IEEEPARstart{N}{etwork} slicing is a critical technology for enabling flexible and intelligent radio access networks (RAN) in next-generation wireless networks, as it allows efficient support for highly dynamic and heterogeneous service demands such as enhanced Mobile Broadband (eMBB), Ultra-Reliable Low-Latency Communications (URLLC), and massive Machine-Type Communications (mMTC)~\cite{oran_wg3_2024}.

Open RAN’s (O-RAN) disaggregated architecture—with its clear separation of control and processing elements like the RAN Intelligent Controller (RIC), Centralized Unit (CU), Distributed Unit (DU), and Radio Unit (RU), opens the door to designing scalable, decentralized intelligence. 
This modular structure is ideal for realizing multi-agent deep reinforcement learning (DRL) systems, where autonomous agents distributed across CU and DU components collaboratively manage the network, coordinated by RIC-based xApps~\cite{alam2024comprehensive,3gppRe18,polese2022understanding,lotfi2022evolutionary,lotfi2024open,lotfiattention,lotfi2024metareinforcementlearningapproach}. 
At the DU level, decentralized agents handle localized control tasks such as resource block (RB) allocation and user scheduling. By coordinating across agents, the system can achieve broader, end-to-end optimization goals. Multi-Agent Reinforcement Learning (MARL) aligns naturally with the modular and distributed nature of O-RAN, supporting adaptive and near real-time decision-making for tasks like dynamic slicing and resource provisioning~\cite{zhang2022federated,naderializadeh2021resource,lotfiattention,lotfi2024metareinforcementlearningapproach}. 
However, standard DRL methods face notable limitations in highly dynamic network environments. They typically demand extensive interaction cycles to learn stable policies and often fail to adapt quickly when network states shift. Additionally, the unstructured, noisy, and delayed feedback inherent to wireless systems makes it challenging for agents to extract robust representations and act effectively. While recent approaches have introduced DRL-based adaptive slicing strategies in O-RAN~\cite{lotfi2022evolutionary, lotfi2024open,lotfiattention, lotfi2024metareinforcementlearningapproach}, most still rely on offline training or static models, limiting their generalization to evolving conditions.

Recent studies have explored integrating Large Language Models (LLMs) into wireless systems to enhance deep learning-based decision-making~\cite{merouane2024large,bariah2023understanding,chen2024communication,lotfi2025llm,melike2024llm,zhang2023controlling,wang2024llm}. While traditional fine-tuning of LLMs enables generalization, it remains resource-intensive, posing challenges for dynamic, resource-constrained environments like O-RAN slicing. To address this, parameter-efficient tuning methods such as LoRA~\cite{hu2022lora} and adapters~\cite{houlsby2019parameter} have emerged, enabling task adaptation with minimal computational overhead. 
%
Managing O-RAN functions such as slicing, scheduling, and dynamic resource allocation requires both domain-specific expertise and adaptive decision-making. Prior approaches have explored either static prompt-based GPT models or domain-specific LLMs like ORANSight in isolation. However, these methods often suffer from poor generalization, limited scalability, and weak knowledge reuse challenges further exacerbated in wireless edge environments with constrained resources.

{Recent work, including \cite{lotfi2025llm}, has explored the potential of integrating large language models (LLMs) into reinforcement learning (RL) pipelines to enhance decision-making in wireless systems. These approaches demonstrate that LLMs can encode semantic knowledge about network behavior, configuration policies, and task priors, which can be leveraged to guide RL agents more effectively in complex environments such as O-RAN. However, we observed that the effectiveness of this integration depends heavily on the relevance and structure of the language inputs. In particular, prompts that are closely aligned with the RL task and environmental state offer significantly greater utility than generic or static textual inputs. 
Building on this insight, we propose a prompt-driven LLM-RL framework in which lightweight, learnable tokens are trained jointly with the RL agent to generate task-specific prompts and adapt a frozen LLM to network slicing and scheduling tasks. 
While this approach reduces training overhead and improves sample efficiency, deploying LLMs directly on edge devices introduces challenges related to memory, latency, and inference cost. 
To address these challenges, we design \textit{ORAN-GUIDE}, a prompt-augmented multi-agent LLM framework inspired by retrieval-augmented generation (RAG) architectures. The design  decouples domain knowledge generation from task execution: \textit{ORANSight}, a fine-tuned domain expert trained on O-RAN documentation and telemetry~\cite{gajjar2025ORANSight}, dynamically generates context-rich prompts based on the current network state. These prompts are consumed by \textit{GPT}, a lightweight general-purpose LLM augmented with learnable prompt tokens,~\cite{radford2019language} which combines the retrieved domain context with semantic representations of the environment to inform RL decision-making. 
To enable practical deployment in edge-based systems, we introduce a \textit{ knowledge distillation (KD) mechanism} that transfers domain-specific insights from ORANSight into the GPT module through task-adaptive prompt tuning. This architecture preserves the rich domain understanding of large models while maintaining a low-parameter, modular design. The result is a scalable and interpretable LLM-augmented RL framework that bridges semantic-level reasoning with low-level control, offering both theoretical flexibility and real-world deployability.} 

\begin{figure*}[t]
  \centering
\includegraphics[width=1.4\columnwidth]{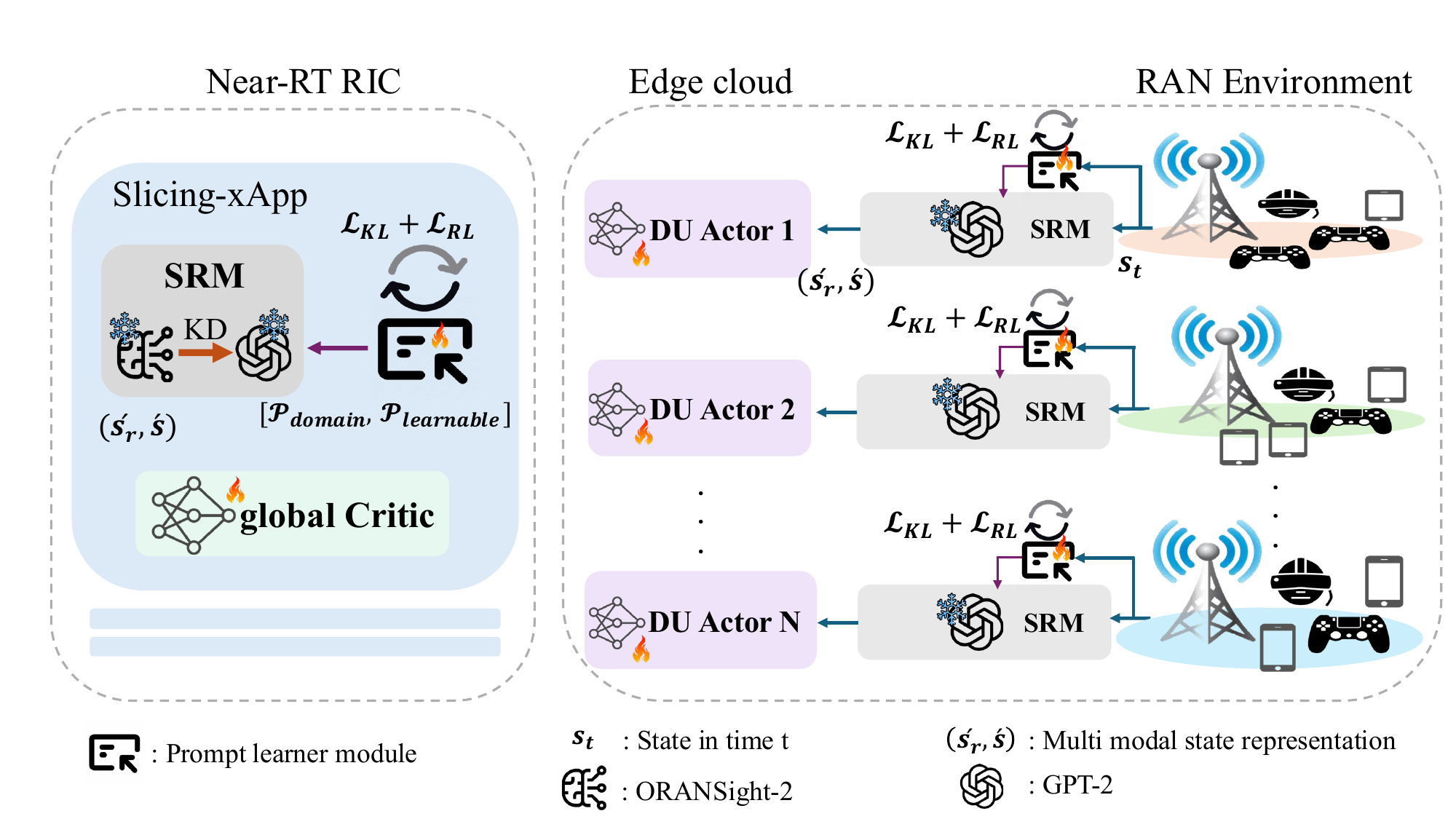}\vspace{-0.2cm}
    \caption{\small Overview of the ORAN-GUIDE system architecture showing distributed DRL agents coordinated by a centralized critic, with prompt-augmented State Representation Modules (SRMs).}\vspace{-0.2cm}
    \label{sys_graph}
\end{figure*}
By decoupling domain knowledge generation from task execution, ORAN-GUIDE reduces the functional burden on each model and enhances system interpretability and modularity. This separation simplifies maintenance and enables flexible updates, where components can be independently retrained or reused. Notably, such modularity supports cross-domain transferability, a limitation of conventional GPT+Prompt or ORANSight+Prompt configurations, which tightly couple domain knowledge with inference. Furthermore, incorporating structured, context-rich prompts into GPT's learning pipeline improves sample efficiency and accelerates convergence by providing task-relevant guidance aligned with O-RAN semantics. Through the synergy of ORANSight’s embedded expertise and GPT’s general reasoning capabilities, the framework achieves stronger generalization across dynamic and heterogeneous wireless environments, offering a scalable solution for LLM-augmented reinforcement learning in specialized domains.

\begin{table}[ht]
\centering
\small 
\caption{\small Comparison of LLM-Based Decision-Making Frameworks in O-RAN Environments}
\label{tab:llm_comparison}
\resizebox{\columnwidth}{!}{%
\begin{tabular}{|p{3.cm}|p{2.cm}|p{2.4cm}|p{2.8cm}|}
\hline
\textbf{Feature} & \textbf{GPT + Static Prompt~\cite{lotfi2025llm}} & \textbf{ORANSight~\cite{gajjar2025ORANSight} + Static Prompt} & \textbf{ORAN-GUIDE (Ours)} \\
\hline
Domain Knowledge Integration & Manual and static & Domain-specific but static & Dynamic, context-aware via ORANSight \\
\hline
Task Adaptability & Low & Medium & High (learnable tokens + GPT inference) \\
\hline
Modularity & No & Limited & Yes (separated knowledge generation and policy) \\
\hline
Cross-Domain Transferability & Poor & Poor & Strong (modular and reusable components) \\
\hline
Sample Efficiency & Low & Moderate & High (prompt-guided learning) \\
\hline
Generalization to New Scenarios & Limited & Limited & Improved (domain expertise + GPT reasoning) \\
\hline
Suitability for Edge Deployment & Inefficient & Inefficient & Efficient (via distillation + compact prompt design) \\
\hline
Scalability to Complex Environments & Poor & Moderate & High \\
\hline
\end{tabular}%
}
\end{table}
As depicted in Fig. \ref{sys_graph}, this design combines the strengths of both LLM models and offers the following key advantages:

\begin{itemize}
    \item \textbf{Role Specialization for Modular Intelligence:} \\
    By decoupling prompt generation and decision-making, our architecture follows a “separation of concerns” principle. ORANSight encapsulates domain-specific priors, trained on configuration logs, slicing strategies, and O-RAN telemetry, to infer environment-specific cues. GPT then receives these cues, along with trainable prompt tokens, to reason about the task and produce policy outputs. This modular design enables each model to specialize, improving efficiency, interpretability, and maintainability.

    \item \textbf{Adaptive Prompting via Technical Knowledge Injection:} \\
    Unlike static or manually written prompts, ORANSight dynamically generates prompts based on the evolving RL state and network context. These prompts convey technical insights such as:
    \begin{enumerate}[label=\alph*)]
        \item Service-level requirements (e.g., Quality of Service (QoS) and latency),
        \item Temporal patterns in resource demand,
        \item Inter-slice interference and congestion.
    \end{enumerate}
    This real-time knowledge injection enhances convergence and robustness, especially in partially observable settings.

\item \textbf{Generalization Across Tasks and Domains:} \\
In practical deployments, O-RAN configurations may vary across domains (e.g., vehicular networks, 6G environments). Our modular setup allows:
\begin{enumerate}[label=\alph*)]
    \item {GPT and its learnable prompt tokens to be adapted or fine-tuned for task-specific behavior,
    \item ORANSight to remain fixed or replaced with a domain-specific variant, depending on the target deployment.}
\end{enumerate}
This separation supports plug-and-play adaptability, enabling domain transfer without retraining the entire decision pipeline.

    \item \textbf{Prompt Fusion for Enhanced Sample Efficiency:} \\
    Our framework supports dual-prompt fusion, combining:
    \begin{enumerate}[label=\alph*)]
        \item $\mathcal{P}_{\text{domain}}$: Context-rich prompts generated by ORANSight,
        \item $\mathcal{P}_{\text{learnable}}$: Task-adaptive prompts tuned via backpropagation during RL training.
    \end{enumerate}
    The fusion of static domain knowledge with dynamic adaptation enables sample-efficient learning and improves policy exploration.

    \item \textbf{Multi-Agent LLM Collaboration as Cognitive Architecture:} \\
    The ORAN-GUIDE setup reflects a form of distributed cognition in AI:
    \begin{enumerate}[label=\alph*)]
        \item ORANSight serves as a “planner” or domain expert,
        \item GPT acts as the “controller” or task executor.
    \end{enumerate}
\end{itemize}

The result is a RAG-inspired decision-making framework, tailored for DRL in technical wireless environments. It leverages a domain-specialized LLM (ORANSight), trained on O-RAN-specific knowledge, to generate contextual prompts that guide policy learning. This structured pipeline, spanning from RAG to LLM-based reasoning and finally to RL-based control, targets critical tasks such as network slicing and RB allocation. By explicitly connecting semantic-level domain understanding with low-level control decisions, the framework offers a unified approach to intelligent decision-making in complex wireless systems.

\emph{To the best of our knowledge, this is the first work to propose a modular, RAG-inspired multi-agent framework that adaptively integrates LLM-based context-aware state representations into DRL for joint O-RAN slicing and scheduling}.  
The paper is structured as follows: Section \ref{sec:literature} reviews related work. Section \ref{sec:system} presents the system model and formulates the joint slicing and scheduling problem in O-RAN. Our proposed RAG-inspired prompt-augmented multi-agent DRL approach is introduced in Section \ref{sec:MHDRL}. Section \ref{sec:simulation} discusses evaluation results and conclusions are given in Section \ref{sec:conclusion}. 

\begin{table*}[ht]
	\footnotesize
	\centering
\caption{Comparison of RAG and LLM-based RL Integration Methods in Wireless Communication Networks}
\begin{tabular}{|p{1cm}|p{2.8cm}|p{2.8cm}|p{1cm}|p{4.5cm}|p{1.2cm}|} 
\hline
\textbf{Ref.} & \textbf{RAG type} & \textbf{LLM type} & \textbf{RL Alg.} & 
\textbf{Objective KPI} & \textbf{O-RAN domain} \\
\hline
\cite{10531073} &  structured knowledge base (KB) retrieval & Custom LLM + ChatGPT & DDPG  &  QoE optimization, latency reduction& $\times$ \\ 
\hline
\cite{huang2024toward} &  semantic RAG framework & Transformer-based & MARL & decentralized orchestration & $\times$ \\ 
\hline
\cite{ren2024retrieval} &  prompt and structured RAG & LLM-enabled IR & DQN &  MEC offloading efficiency, resource utilization & $\times$ \\ 
\hline
\cite{nazar2024enwar} &  multi-modal RAG & Multimodal LLM & PPO &  mobility adaptation & $\times$ \\ 
\hline
\cite{shokrnezhad2025autonomous} &  tokenized API response and RAG & BERT-style encoder LLM & AC &  Dynamic orchestration& $\times$ \\ 
\hline
\cite{shao2024wirelessllm} & Partial (prompt-enhanced RAG) & Prompt-LLMs (GPT \& others) & RLHF  & Prompt relevance, inference accuracy, adaptability & $\times$ \\ 
\hline
\cite{yilma2025telecomrag} &  Knowledge Graph + Document Retrieval & LLama + GPT & RLHF & Protocol compliance, semantic similarity & $\times$ \\ 
\hline
\cite{lotfi2llm}& $\times$ &  GPT-2 &  SAC & network slicing and scheduling& $\checkmark$ \\
\hline
\cite{gajjar2025ORANSight,gajjar2024oran}& $\checkmark$ &  ORANSight & $\times$ & $\times$ & $\checkmark$ \\
\hline
\textbf{This Work} &  custom, structured RAG using ORANSight & ORANSight  & SAC &  QoS-aware network slicing and adaptive RB allocation& $\checkmark$ \\ 
\hline
\end{tabular}
\label{tab:rag_llm_rl_comparison}
\end{table*}

\section{Related Works}\label{sec:literature}


Recent developments in wireless communication have drawn increasing attention to the potential of LLMs as a foundational component in intelligent network systems. Their ability to reason over heterogeneous data modalities, including textual descriptions, protocol specifications, and RF signal measurements, has enabled new capabilities in network management and the automated interpretation of 3GPP standards.~\cite{lotfi2025llm,melike2024llm,gajjar2025ORANSight, chen2024communication, zhang2023controlling,wang2024llm}.

\subsection{Transfer Learning Foundations}

Transfer learning is a core technique in deep learning that helps models apply knowledge learned from large-scale text datasets to new tasks. It usually involves two main steps: first, the model is pretrained on general data; then, it’s fine-tuned using task-specific examples. Although this process improves adaptability, fine-tuning can be expensive in terms of computation and requires a lot of labeled data, making it less practical for dynamic and resource, limited environments like O-RAN slicing. 
To address these challenges, recent studies have proposed more efficient alternatives such as LoRA (Low-Rank Adaptation)\cite{hu2022lora} and adapter-based learning\cite{houlsby2019parameter}. These methods add small trainable components to a frozen pre-trained model, helping it learn new tasks with much lower memory and compute requirements. 
Another promising solution is prompt learning~\cite{lester2021power,li2021prefix}, which keeps the model itself unchanged and instead trains short input prompts to guide its behavior. This approach is particularly useful in few-shot or zero-shot settings, where labeled data is limited, and has shown strong potential for lightweight, flexible adaptation.

\subsection{LLM-Augmented RL in Wireless Network Settings}
Several works have explored LLMs for enhancing RL agents through instruction tuning, reward shaping, and policy guidance through semantic understanding. However, these studies typically rely on static prompts or fixed knowledge bases, limiting adaptability to dynamic environments like wireless networks. 
While prompt learning has gained traction in NLP, recent work is beginning to explore how these ideas, when combined with reinforcement learning, can support dynamic decision-making in wireless environments. 
For instance,\cite{chen2024communication} proposes an offline framework where LLMs assist with interpreting environmental signals and generating feedback in networked control systems. While effective in structured scenarios, the framework lacks real-time responsiveness and struggles in dynamic infrastructures like O-RAN due to its reliance on static training data. In a different approach,\cite{melike2024llm} combines hierarchical RL with LLMs to map operator intent into network actions through rApp/xApp selection. Although this method shows improvements in key performance metrics such as latency and throughput, it is limited by predefined intent templates and reacts primarily to immediate network changes without anticipating future demands. 
Other works have examined LLMs within multi-agent RL contexts. For example,\cite{zhang2023controlling} introduces a centralized critic with token-based feedback for cooperative tasks like resource allocation, and\cite{wang2024llm} enhances learning efficiency by generating contextual state embeddings and intrinsic rewards using LLMs. Despite their contributions, both approaches rely heavily on centralized decision architectures and static knowledge encoding, which restrict their applicability in decentralized, real-time systems like O-RAN.


\subsection{Contextualization and RAG based RL augmented in Wireless Network optimization}

Recent research has explored integrating RAG and LLMs into DRL, particularly for wireless communication and network orchestration. In \cite{10531073}, a structured RAG mechanism generates enhanced observation embeddings for a DDPG agent in next-generation networking. Similarly, \cite{huang2024toward} introduces a semantic RAG framework in 6G systems to inform MARL via environment-aware state representations. In mobile edge computing, \cite{ren2024retrieval} applies prompt-driven RAG outputs embedded by an LLM to guide a deep Q-learning agent for resource allocation. 
Multimodal settings have also been addressed. For instance, \cite{nazar2024enwar} fuses wireless sensor modalities into a perception vector for a PPO agent using a RAG pipeline, and \cite{shokrnezhad2025autonomous} leverages a BERT-style LLM with tokenized RAG responses to support continual RL-based orchestration. Meanwhile, \cite{shao2024wirelessllm} combines RAG-informed prompts with RLHF or policy-gradient methods, though typically relying on general-purpose LLMs like GPT or LLaMA, with RAG treated more as a preprocessing tool than a deeply integrated module. Even domain-specific studies such as \cite{yilma2025telecomrag}, which retrieve telecom protocol documents, often exhibit loose coupling between RAG, LLM, and DRL, lacking semantic alignment with real-time constraints.

In the O-RAN domain, \cite{lotfi2llm} integrates a general-purpose LLM with DRL for dynamic network slicing, demonstrating natural language reasoning's potential in policy learning. ORANSight~\cite{gajjar2025ORANSight}, a domain-specific LLM with RAG mechanisms, contextualizes O-RAN knowledge, while Oran-Bench-13k~\cite{gajjar2024oran} offers a benchmark to evaluate RAG-based LLM performance on O-RAN tasks. These works highlight individual components of the LLM-DRL-RAG paradigm but remain modular or exploratory, without forming a unified, closed-loop control system. 
In contrast, our work introduces a fully integrated architecture that combines structured RAG, a domain-trained LLM (ORANSight), and DRL to support closed-loop control in O-RAN. Structured telemetry data—such as QoS targets, SNR, and RB usage is processed through a task-aware RAG module and interpreted by ORANSight to generate context-rich prompts. These semantic cues are consumed by a Soft Actor-Critic (SAC) agent for adaptive slicing and resource allocation. To further enhance adaptability and efficiency, we propose a prompt-augmented DRL framework that avoids full model fine-tuning. By embedding contextual prompts into the DRL loop, decentralized agents can make dynamic slicing decisions more efficiently. This lightweight and flexible design improves generalization, responsiveness, and scalability—making it well-suited for fast-changing RAN conditions.

\section{SYSTEM MODEL AND PROBLEM FORMULATION}\label{sec:system}
\subsection{O-RAN Slicing Architecture}
We consider a virtualized O-RAN system divided into three functional slices, each tailored to distinct QoS requirements: enhanced Mobile Broadband (eMBB) for high-throughput applications, massive Machine-Type Communications (mMTC) for dense device connectivity, and Ultra-Reliable Low-Latency Communication (URLLC) for ultra-low latency communication. The RIC oversees dynamic resource distribution across network components, CUs, DUs, and RUs, through a hierarchical control strategy. At the upper tier, resources are allocated between slices based on slice-level service demands (inter-slice allocation), while within each slice, scheduling mechanisms distribute capacity among UEs according to slice-specific policies (intra-slice allocation). These decisions are enforced at the physical layer by the MAC protocol. A high-level schematic of this architecture is shown in Fig.~\ref{sys_graph}.


\subsection{UE's Achievable Data Rate}
O-RAN employs Orthogonal Frequency Division Multiplexing (OFDM) to mitigate intra-cell interference~\cite{3gpp15}, and QoS for each slice is characterized through key performance indicators (KPIs). A fundamental KPI is the achievable data rate for each user equipment (UE). 
Each UE’s data rate is influenced by its channel conditions, resource assignments, and interference from neighboring cells. Assuming an OFDM-based air interface, the instantaneous data rate for a UE $u$ associated with slice $l$ at RU $m$ is expressed as:
\begin{equation}
c_{l,u,m}(t) = B \sum_{k=1}^{K_{l,m}} e_{u,k} b_{l,k} \log\left(1 + \frac{p_{k,m}(t) g_{u,k}(t)}{I_{u,k}(t) + \sigma^2}\right),
\end{equation}
where $g_{u,k}(t)$ captures the combined effect of path loss and small-scale fading, $I_{u,k}(t)$ is the interference power from other cells, and $B$ denotes the RB bandwidth. Moreover, $e_{u,k} \in {0,1}$ is a binary variable indicating whether RB $k$ is assigned to UE $u$.
$b_{l,k}$ represents the RB allocation indicator for slice $l$, and $p_{k,m}$ denotes the transmission power. 
$K_{l,m}$ is the number of RBs assigned to slice $l$ at RU $m$. 
$I_{u,k}(t)$ is the inter-cell interference
and $\sigma^2$ is the variance of Additive White Gaussian Noise (AWGN).



\subsection{Slice-Specific QoS Objectives}










Each slice defines its own QoS goals via relevant KPIs. 

\noindent\textbf{Slice 1:} (eMBB-Throughput-Oriented Performance)

The main objective of the eMBB slice is to deliver high data rates across all connected UEs. Accordingly, its QoS is evaluated based on the average aggregate user throughput:
\begin{align} 
\mu_r = \frac{1}{N_u}\sum_{i=1}^{N_u} C_i, 
\end{align}
where $C_i$ denotes the instantaneous data rate of user $i$, and $N_u$ is the total number of users in the slice. This metric captures the system’s ability to deliver consistently high throughput in bandwidth-intensive applications such as video streaming and cloud gaming.

\noindent\textbf{Slice 2:} (mMTC-Scalability and Connectivity)

The mMTC slice targets massive device connectivity with low-to-moderate throughput requirements. Its QoS metric balances connection density and per-device throughput by combining an availability indicator with aggregate performance:
\begin{align} 
d_s = \frac{\sum_{i=1}^{N_u} \mathds{1}{(C_i > \lambda_i)}}{N_u} \sum_{i=1}^{N_u} C_i,
\end{align}
where $\mathds{1}(C_i > \lambda_i)$ is an indicator function that equals 1 if user $i$'s throughput exceeds a minimum threshold $\lambda_i$, and 0 otherwise. This formulation ensures that the system not only supports a large number of devices but also guarantees basic service quality across active users.

\noindent\textbf{Slice 3:} (URLLC-Latency-Sensitive Applications)

For latency-critical services such as industrial control or autonomous driving, minimizing delay is the top priority. The URLLC slice is evaluated based on the worst-case latency experienced among all users:
\begin{align} 
l_d = \max (\tau_i), \quad \forall i \in N_u,
\end{align}
where $\tau_i$ denotes the end-to-end latency for user $i$. This maximum delay metric reflects the slice's ability to maintain strict delay bounds under dynamic conditions.

To enable joint optimization across slices, a unified QoS vector is defined as:
\begin{align}
\boldsymbol{Q} = [\mu_r, d_s, l_d], \quad \text{and} \quad Q_l = \boldsymbol{Q}[l], \quad \forall l \in \{1, 2, 3\},
\end{align}
where $Q_l$ extracts the slice-specific QoS metric from the global QoS vector $\boldsymbol{Q}$. This formulation enables slice-aware optimization in both training and evaluation processes.\vspace{-0.2cm}

\subsection{Problem Formulation: Utility-Aware Resource Optimization}

To balance fairness and performance across diverse O-RAN slices, we define a utility-based optimization framework. Each slice $l$ is associated with a utility function $U_l$, reflecting its QoS level, and weighted by a priority coefficient $w_l$ that encodes its service importance. This formulation ensures that slices with critical demands (e.g., URLLC) receive prioritized treatment, while preserving adaptability and resource fairness system-wide. 
The optimization objective is expressed as:
\begin{subequations}
\begin{align}\label{opt1}
\max_{\boldsymbol{b}, \boldsymbol{e}} \quad & \sum_{l \in \mathcal{L}} w_l U_l(\boldsymbol{b}, \boldsymbol{e}) \\
\text{s.t.} \quad & \boldsymbol{b}_l \in \{0,1\}^{|\mathcal{L}| \times K_{l,m}}, \quad \boldsymbol{e}_l \in \{0,1\}^{N_{u,m} \times K_{l,m}}, \\
& \sum_{l} \sum_{u} \sum_{k} e_{u,k} b_{l,k} \leq K_m, \label{cons:capacity} \\
& \sum_l b_{l,k} \leq 1 + \lambda_l \max(0, \sum_l b_{l,k} - 1), \label{cons:sharing} \\
& Q_{\boldsymbol{b}_l, \boldsymbol{e}_l} \geq Q_{{\min},l} - \delta_l, \quad \forall l. \label{cons:qos}
\end{align}
\end{subequations}
Here, $K_m$ is the available number of resource blocks (RBs) at RU-DU pair $m$, and $N_{u,m}$ represents the number of connected UEs. Constraint \eqref{cons:capacity} enforces resource budget limits, while \eqref{cons:sharing} allows conditional RB sharing using a tunable relaxation term $\lambda_l$ to accommodate temporary overloads. Constraint \eqref{cons:qos} ensures that each slice meets a relaxed QoS requirement to prevent infeasibility under dynamic demand patterns.

Given the time-varying and stochastic nature of the wireless environment, solving this optimization in a static form is impractical. Instead, we formulate the problem as a Markov Decision Process (MDP), where the agent observes system state variables, such as per-slice QoS indicators and resource availability, and makes sequential allocation decisions. 
To learn adaptive and robust slicing strategies under this MDP framework, we adopt DRL techniques. However, due to the multi-agent interactions and non-stationary dynamics, DRL alone can face convergence and generalization challenges. To address this, we introduce a multi-agent LLM prompt-guided DRL scheme, where a domain-specific pretrained LLM collaborate with a lightweight LLM to generate structured prompts and provide contextual guidance for decision-making. This hybrid setup improves policy stability, responsiveness, and interpretability in real-time O-RAN control.\vspace{-0.2cm}





\begin{figure}[t]
  \centering
\includegraphics[width=\columnwidth]{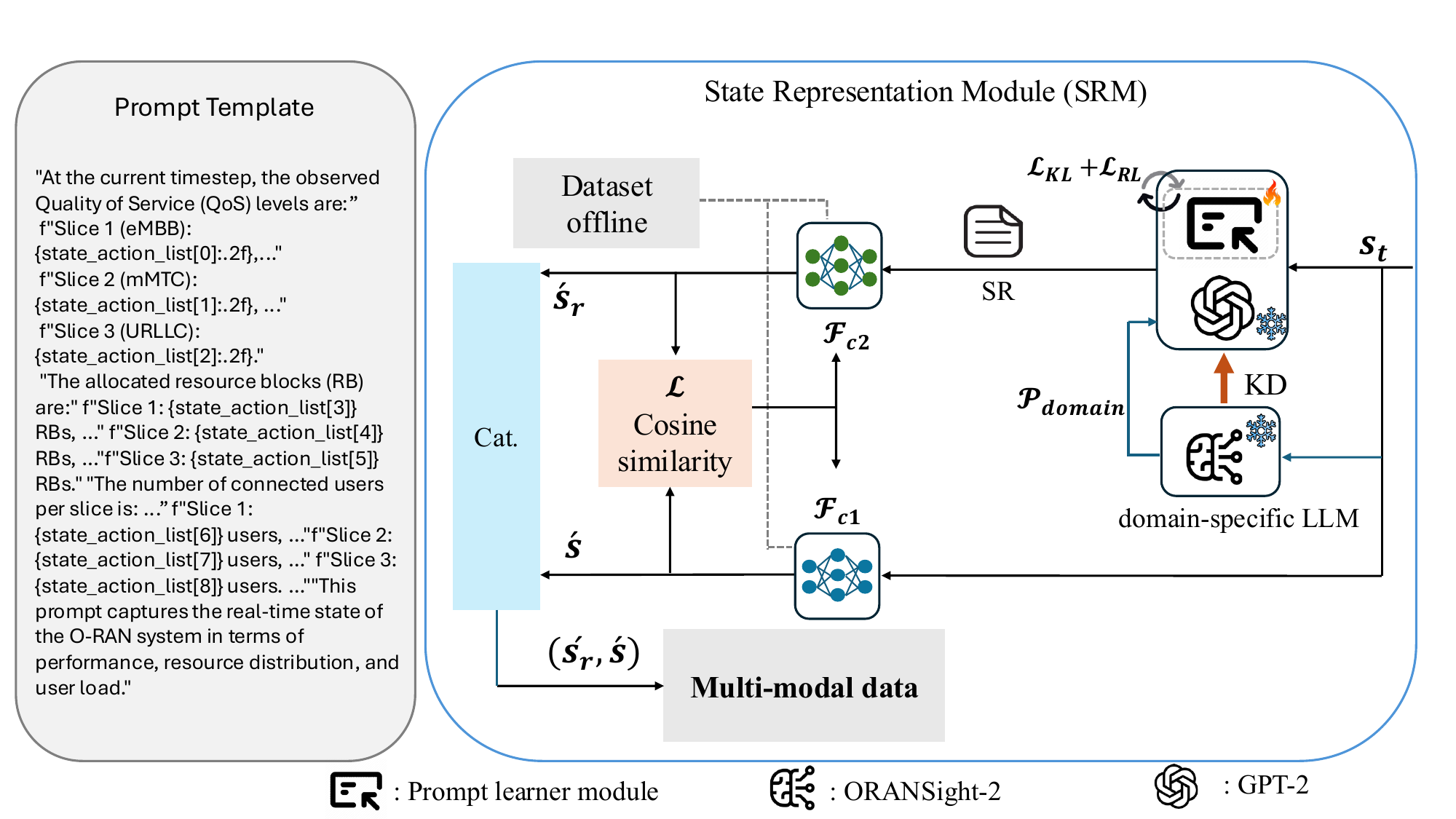}\vspace{-0.2cm}
    \caption{\small SRM design illustrating how domain-specific and learnable prompts are fused and encoded by GPT to produce semantic state embeddings. 
    }\vspace{-0.2cm}
    \label{srm}
\end{figure}
\section{The ORAN-GUIDE Framework
}\label{sec:MHDRL}
\subsection{Overview of the Proposed Architecture}


In this framework, we leverage a domain-specific large language model, ORANSight, to implement a RAG-style prompt generation mechanism tailored for reinforcement learning in O-RAN systems. Rather than relying on static or manually defined prompts, we supply ORANSight with structured, real-time network information, such as slice-level QoS targets, SNR measurements, RB allocations, and traffic trends. ORANSight uses this data to produce contextual prompts that encode domain-relevant insights for the RL agent. 
By embedding this knowledge into the DRL learning process, the agent operates with improved awareness of system priorities and constraints, leading to better decision-making and greater adaptability in dynamic environments. 
The system architecture features a multi-agent LLM design. ORANSight acts as the domain knowledge generator, translating network telemetry into high-level prompts. These are then consumed by a general-purpose LLM, like GPT, equipped with learnable prompt tokens, which uses them to guide policy decisions. This collaboration enables the agent to combine structured knowledge with flexible reasoning-bridging domain understanding and control execution within a unified DRL framework. 
This modular approach not only enhances the agent's performance and generalization but also separates domain knowledge from task execution, making the framework both scalable and adaptable to diverse O-RAN deployment scenarios.\vspace{-0.2cm}

\subsection{MDP Formulation}
To address the dynamic optimization problem in Eq.~\eqref{opt1}, we model the system as a MDP, defined by the tuple $\langle \mathcal{S}, \mathcal{A}, T, \gamma, r \rangle$. This formulation allows an RL agent to iteratively learn an optimal policy for resource allocation under changing network conditions. Here, $T$ denotes the transition function governing the probability distribution of reaching a new state $s_{t+1}$ given the current state $s_t$ and action $a_t$. The discount factor $\gamma \in [0,1]$ is used to balance immediate and future rewards. 
The components of the MDP are detailed below:

\textbf{(1) State Space ($\mathcal{S}$):}
At each time step $t$, the environment is described by a state vector $s_t \in \mathcal{S}$, which captures key network information required for decision-making. This includes current slice-specific QoS levels ($Q_l$), the number of active UEs per slice ($N_{l,u}$), and the previous allocation decision $a_{t-1}$. Formally, expressed as $s_t = \{ Q_l, N_{l,u}, a_{t-1} \mid \forall l \in \mathcal{L} \}$.

\textbf{(2) Action Space ($\mathcal{A}$):}
The agent’s action at time $t$ is denoted by $a_t \in \mathcal{A}$ and reflects a complete resource allocation decision. This includes both inter-slice bandwidth partitioning ($\boldsymbol{b}$) and intra-slice resource scheduling ($\boldsymbol{e}$), according to $a_t = \{ \boldsymbol{b}, \boldsymbol{e} \}$.

\textbf{(3) Reward Function ($r$):}

To guide the agent toward fair and effective resource allocation, we define a reward function that unifies heterogeneous QoS objectives under a single maximization goal. Since some metrics, like latency, are minimization-based, we invert them relative to a reference value, while keeping throughput-based metrics unchanged. Each slice's intermediate reward $r_{0,l}$ is computed using a piecewise function:
\begin{equation}
r_{0,l} =
\begin{cases}
1 + \beta \cdot \frac{\overline{Q}_l - \text{thr}}{\text{thr}}, & \overline{Q}_l \geq \text{thr}, \\
\exp\left(-\gamma \cdot \frac{\text{thr} - \overline{Q}_l}{\text{thr}}\right), & \text{otherwise}.
\end{cases}
\end{equation}
These values are normalized via sigmoid smoothing:
\begin{equation}
r_Q = \sum_l \frac{1}{1 + \exp(-\alpha \cdot r_{0,l})},
\end{equation}
and penalized for severe underperformance as negative reward
\begin{equation}
r_{ng} = \sum_{l: \overline{Q}_l < \text{thr}(1 - \text{margin})} \exp\left(-\delta \cdot \frac{\overline{Q}_l - \text{thr}}{\text{thr}}\right).
\end{equation}
The final reward expressed as $r_t = r_Q - r_{ng}$.

The agent learns a policy $\pi(a_t \mid s_t)$ to maximize expected returns $R(t) = \sum_{i=0}^\infty \gamma^i r_{t+i}$, using standard value functions $V(s_t)$ and $Q(s_t, a_t)$ for policy evaluation and improvement.\vspace{-0.2cm}
\subsection{Prompt-Augmented DRL Architecture}
\begin{figure}[t!]
  \centering
\includegraphics[width=\columnwidth]{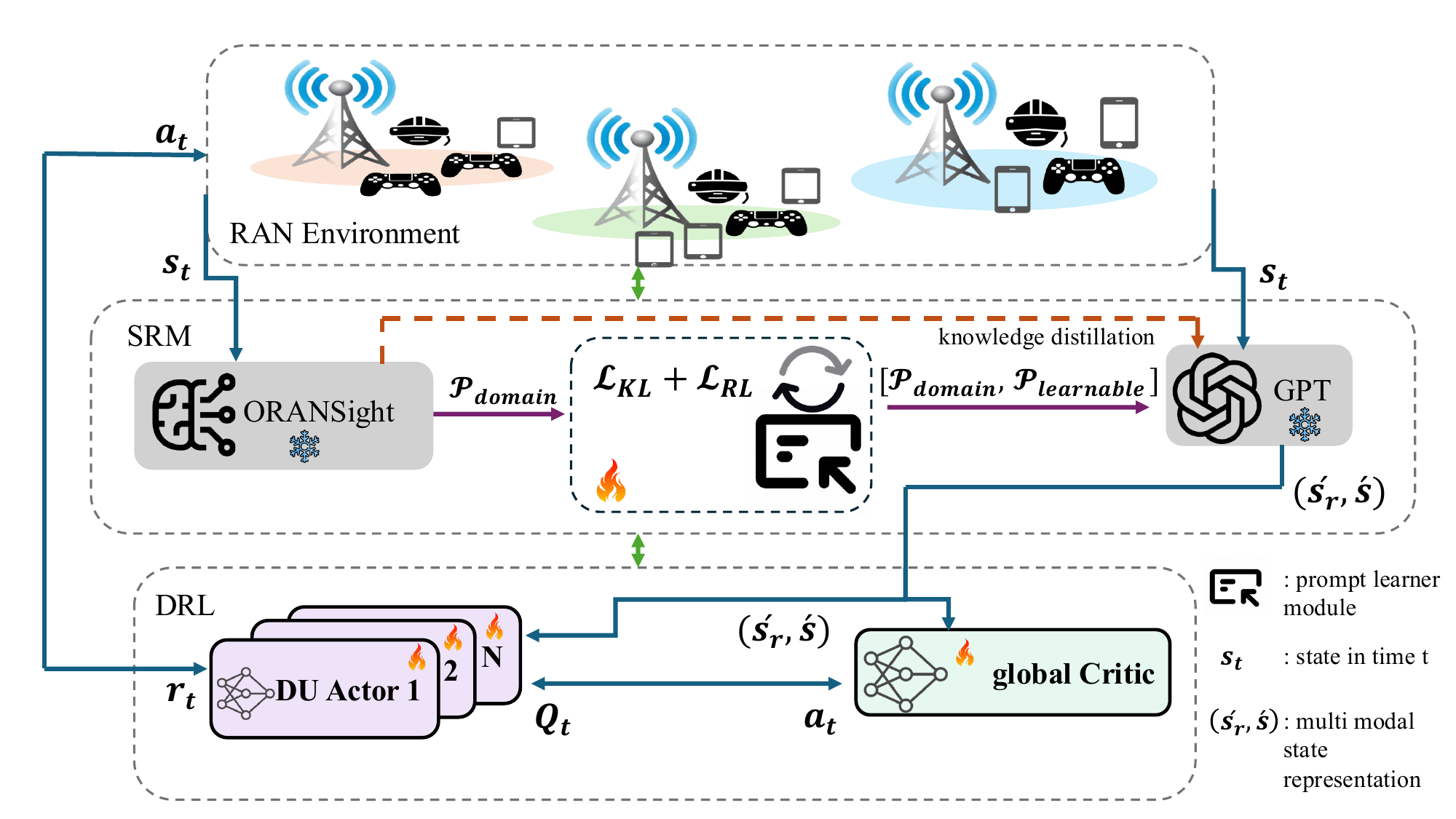}\vspace{-0.2cm}
    \caption{\small Training workflow of ORAN-GUIDE, integrating dual LLM prompts into DRL via a soft actor-critic loop, with distillation and prompt fusion for adaptive policy learning.}\vspace{-0.2cm}
    \label{train_GUIDE}
\end{figure}

Building on our prior work~\cite{lotfi2025llm}, which employed informal fixed prompts to convert network parameters into structured textual inputs for LLM-guided decision-making, we extend this framework by introducing domain-aware and learnable prompts. These prompts enhance the model’s adaptability and improve its ability to extract and represent semantic patterns relevant to dynamic O-RAN scenarios. 
To effectively address the formulated MDP and improve policy learning under uncertainty, we incorporate LLMs into the decision-making pipeline of DRL framework. Conventional DRL agents often struggle with sample inefficiency and poor generalization challenges that are amplified in high-dimensional and dynamic environments such as O-RAN slicing. 
To overcome these limitations, ORAN-GUIDE integrates LLM-driven prompting into the state representation module (SRM), depicted in Fig.~\ref{srm}, enabling the DRL agent to operate with enriched semantic context and domain-aligned cues. Our architecture leverages two complementary prompting mechanisms:
\begin{enumerate}[left=0pt]
\item \textbf{Domain-Aware Prompts ($\mathcal{P}_{\text{domain}}$)}: Generated by the ORANSight model~\cite{gajjar2025ORANSight}, these prompts encode O-RAN-specific insights such as service-level objectives, traffic trends, and interference conditions. They provide scenario-aware guidance that enhances decision flexibility and contextual awareness.
\item \textbf{Learnable Prompts ($\mathcal{P}_{\text{learnable}}$)}: These are trainable prompt tokens that embed task-specific knowledge and evolve during DRL training. By interacting with GPT’s frozen backbone, they help shape policy representations, improving both stability and generalization across variable network states.
\end{enumerate}
Together, these prompts are fused into a single input stream that drives the SRM. This prompt-driven adaptation enhances the agent’s ability to learn robust and sample-efficient policies for real-time network slicing and resource allocation. 
The high-level interaction between the DRL agent, SRM, and RAN environment during training process is illustrated in Fig.~\ref{train_GUIDE}, while Fig.~\ref{sys_graph} shows high-level overview of ORAN-GUIDE framework, and the Fig.~\ref{srm} is detailed internal structure of the SRM, which handles prompt fusion and semantic encoding.
\subsection{ORANSight: The Domain-Aware Prompt Generator}
ORANSight is a lightweight LLM fine-tuned on O-RAN-specific data, including slicing policies, configuration logs, and control signals. Its role is to interpret the current network state and generate technical prompts that help the decision-making agent act more effectively. 
At each time step $t$, ORANSight receives a structured snapshot of the environment as $s_t$. Based on this input, it produces a domain prompt $\mathcal{P}_{\text{domain}}$ that summarizes key context, highlights critical conditions (e.g., congestion or QoS violations), and provides useful guidance to the DRL agent. 
Unlike fixed or manually written prompts, ORANSight generates dynamic, real-time prompts tailored to the system’s current state. 
This mechanism follows a Retrieval-Augmented Generation (RAG) paradigm, where outputs are enriched using information retrieved from relevant domain knowledge. In our case, ORANSight is a domain-specific LLM, pretrained on O-RAN corpora such as slicing policies and configuration logs. At runtime, it performs dynamic RAG by transforming real-time telemetry and network conditions into structured, context-aware prompts. This bridges learned system knowledge with live observations, guiding the RL agent toward more adaptive and informed decisions. \vspace{-0.2cm}




\subsection{GPT with Learnable Prompts} 
The general purpose GPT model~\cite{radford2019language} in our framework acts as a semantic encoder that transforms structured domain knowledge and environmental observations into high-level representations for DRL agent. It receives $\mathcal{P}_{\text{domain}}$ and $\mathcal{P}_{\text{learnable}}$ tokens as input. 
These components are embedded into a single token sequence and passed through the frozen GPT model. Unlike ORANSight, which encodes external knowledge, the learnable prompts capture task-specific inductive biases and are updated during DRL training loop via gradient descent. 
Rather than directly predicting actions or value functions, GPT produces a semantic embedding of the current state, enriched with domain context. This embedding is then used by the downstream DRL agent (e.g., actor or critic network) to make resource allocation decisions. In essence, GPT serves as a prompt-driven state encoder, enabling the agent to reason over both learned task dynamics and structured, language-based cues. 
By combining domain-specific prompting with in-task adaptation, this design enables robust, interpretable, and sample-efficient policy learning in complex O-RAN environments.
\subsection{Prompt Fusion and Policy Learning}
The combined token sequence $[\mathcal{P}_{\text{domain}}(s_t), \mathcal{P}_{\text{learnable}}]$ is processed by the GPT model to produce a context-enriched semantic representation of the current environment. This dual-prompt fusion strategy enables the model to integrate both static domain knowledge (via ORANSight) and dynamic task-specific adaptation (via learnable prompts). 
To align the semantic outputs of the LLM with numerical RL inputs, we incorporate two lightweight adapter networks, $\mathcal{F}{c_1}$ and $\mathcal{F}{c_2}$, trained offline using paired data from real environment states $s_t$ and their corresponding LLM-generated representations $s_{r,t}$. These adapters, adapted from our previous work~\cite{lotfi2025llm}, map both modalities into a unified latent space suitable for downstream policy learning. This design enables cohesive processing of hybrid state features, facilitating smoother integration between textual and numeric state components. Policy learning is performed through a RL loop (e.g., SAC), where the reward signal guides the optimization of the learnable prompts while the GPT backbone remains frozen. To support lightweight edge deployment, we further introduce a KD mechanism: knowledge from the domain-specific ORANSight model is distilled into the general-purpose GPT agent via prompt tuning. This allows the learnable prompts to internalize domain-specific insights without requiring full LLM fine-tuning. 
As a result, the distilled GPT model, along with its adapter-enhanced SRM, can operate independently at the network edge, reducing reliance on centralized inference and enabling scalable, low-latency deployment.
\subsection{DRL Training Loop with SAC Updates}
To optimize the policy network in ORAN-GUIDE framework, we adopt the Soft Actor-Critic (SAC) algorithm within a MARL setup. SAC is well-suited for the O-RAN domain due to its support for continuous action spaces and entropy-based exploration. 
In our architecture, distributed SAC actor agents operate locally at DUs, while a centralized critic in the near-RT RIC performs coordinated value estimation and policy updates. Each agent receives a fused input sequence comprising domain prompts, learnable prompts, and environment states. This sequence is processed by the GPT-based encoder to produce a semantic representation, which guides both action selection and value estimation. 
Training proceeds by collecting transitions, updating the critic using temporal-difference loss, and optimizing the actor using the entropy-regularized policy gradient:
\begin{align}\label{actor_update}
\nabla_{\theta_p}J(\pi_{\theta_p}) &= \mathbb{E}{\kappa,\pi} \Big[ \nabla{\theta_p} \log \pi_{\theta_p}(a_t | s_t) \cdot
\big( Q_v(s_t, a_t) \nonumber\\
&- \beta \log \pi_{\theta_p}(a_t | s_t) \big) \Big], 
\end{align}
where $\beta$ controls the balance between reward maximization and entropy regularization. The Q-value function is defined as $Q_v(s_t, a_t) = \mathbb{E}_{\pi}[R(t) \mid s_t, a_t]$. The critic and the value network parameters $\theta_v$ are updated by minimizing:
\begin{align}\label{critic_update}
\min_{\theta_v} \mathbb{E}_{\kappa,\pi} \left[\left(y_t - Q_v(s_t, a_t)\right)^2 \right],
\end{align}
where the target $ y_t = r_t + \gamma Q_v(s_{t+1}, a_{t+1}; \theta_v) - \beta \log \pi_{\theta_p}(a_{t+1} | s_{t+1})$ incorporates the next state’s Q-value and policy entropy. 
During this loop, the learnable prompts are optimized via gradients from the DRL loss, and KD from the ORANSight model further refines them by aligning the general-purpose GPT’s representations with domain-specific insights. This fusion of symbolic guidance and policy learning enhances both adaptability and generalization across network conditions.

\subsection{ORAN-GUIDE Network Slicing Algorithm} 
The proposed approach, summarized in Algorithm~\ref{alg:oran-guide}, outlines the training loop of the ORAN-GUIDE framework for adaptive network slicing in O-RAN environments. The algorithm combines a MARL setup with LLM integration through dual prompting mechanisms. 
At each iteration $t$, structured network state information $s_t$ is used by the domain-specific LLM (ORANSight) $\mathcal{M}_d$ to generate context-aware prompts $\mathcal{P}_{\text{domain}}$. These are fused with a set of learnable prompts $\mathcal{P}_{\text{learnable}}$ and passed to a frozen GPT model $\mathcal{M}_g$, which produces a semantic representation of the current environment. This embedding guides policy $\theta_p$ and value $\theta_v$ updates using the SAC algorithm. 
To improve generalization and enable lightweight edge deployment, a KD step transfers insights from ORANSight $\mathcal{M}_d(s_t)$ to the learnable prompt module $\mathcal{M}_g(\mathcal{X}_t)$. This design allows for domain-aligned, sample-efficient policy learning while maintaining modularity and scalability across distributed O-RAN agents. 


\begin{algorithm}[t!]
\SetAlgoLined
\textbf{Input}: Number of iterations $N_t$, actors $N_m$, evaluations $N_e$, actor weights $\theta_{p,i}$, critic weights $\theta_v$, \\
\phantom{\textbf{Input}:} ORANSight model $\mathcal{M}_d$, general GPT model $\mathcal{M}_g$, $\mathcal{P}_{\text{learnable}}$,\\
\phantom{\textbf{Input}:} adapter networks $\mathcal{F}_{c1}, \mathcal{F}_{c2}$, distillation coefficient $\lambda$ \\
\textbf{Initialize:} $\theta_{p,i}$, $\theta_v$, and $\mathcal{P}_{\text{learnable}}$ \\

\For{iteration $t=1:N_t$}{
    \For{actor $i=1:N_m$}{
        Obtain state $s_t$ from environment \\
        $\mathcal{P}_{\text{domain}}(s_t) \gets \mathcal{M}_d(s_t)$ \tcp*{ORANSight generates domain prompt} 
        Token sequence $\mathcal{X}_t \gets [\mathcal{P}_{\text{domain}}(s_t), \mathcal{P}_{\text{learnable}}]$ \\
        $h_t \gets \mathcal{M}_g(\mathcal{X}_t)$ \tcp*{GPT outputs prompt-fused embedding}
        
        $s'_{r,t} \gets \mathcal{F}_{c2}(h_t)$, $s'_t \gets \mathcal{F}_{c1}(s_t)$ \\
        $r_i = \text{evaluate}(\pi_{p,i}(s'_{r,t}, s'_t))$ \\
        Store transition $\mathcal{B} \gets \langle (s'_{r,t}, s'_t), a_t, (s'_{r,t+1}, s'_{t+1}), r_t \rangle$ \\

        Compute $\mathcal{L}_{\text{distill}} = \text{KL}(\mathcal{M}_g(\mathcal{X}_t) \Vert \mathcal{M}_d(s_t))$ \tcp*{Distill domain knowledge}
    }
}
Update critic network $\theta_v$ using \eqref{critic_update}. \\
Update actor networks $\theta_{p,i}$ using \eqref{actor_update}. \\
Update prompt embeddings $\mathcal{P}_{\text{learnable}}$ via combined loss: 
$\mathcal{L}_{\text{total}} = \mathcal{L}_{\text{RL}} + \lambda \cdot \mathcal{L}_{\text{distill}}$ \\

\If{$\theta_{p, \forall i}$ converges}{
    Break
}
\textbf{Output}: Trained actor weights $\theta_{p,i}$, critic weights $\theta_v$, optimized prompt embeddings $\mathcal{P}_{\text{learnable}}$ \\
\caption{ORAN-GUIDE: Prompt-Fused RL with Dual-LLM Distillation}
\label{alg:oran-guide}
\end{algorithm}

% ...


% ...





\section{Evaluation Results}\label{sec:simulation}
\subsection{Simulator and Parameter Settings}
We evaluate our approach in a simulated O-RAN environment consisting of three slices, eMBB, mMTC, and URLLC that distributed across $N_m = 6$ DUs. Each DU operates as an autonomous agent responsible for managing local traffic dynamics and service requirements. The user set ($N_u = {50, 100, 200}$) is initialized with uniform spatial distribution, and user mobility follows the direction model proposed in~\cite{lotfi2025llm}, where velocities range from $10$ to $20$ m/s and movement occurs along one of seven discrete angles ${0, \pm \pi/12, \pm \pi/6, \pm \pi/3}$. The network setup assumes Rayleigh fading, AWGN, and adopts a standard 3GPP-compliant PHY layer with subcarrier spacing of $15$ kHz, total DU bandwidth of $20$ MHz, RB size of $200$ kHz, and user transmit power set to $p_u = 56$ dBm, according to~\cite{3gpp15}.

Our algorithm is implemented in PyTorch using an actor-critic setup, where both actor and critic networks contain three fully connected layers of sizes $600$, $700$, $700$ and use \textit{Relu} activations. Training is conducted with the Adam optimizer using a learning rate of $1e-4$ and a batch size of $128$. 
We compare three variants of our method: (1) the proposed ORAN-GUIDE, a multi agent LLM models with integerating a domain-trained ORANSight-2.0~\cite{gajjar2025ORANSight} based on Mistral AI 7B, and general model (GPT-2~\cite{radford2019language}) with learnable prompt tokens to provide semantically enriched state inputs for both actor and critic networks; (2) ORANSight prompt augmented MARL (ORANSight PA-MARL), which integrates a domain-specific LLM with prompt-aligned SRMs to provide context-aware insights at both the actor and critic levels; (3) GPT prompt augmented MARL (GPT PA-MARL), which replaces ORANSight with GPT-2 while retaining prompt alignment; and (4) ORANSight MARL, which uses the ORANSight LLM within the SRM but omits prompt alignment~\cite{lotfi2025llm}. The standard MARL baseline is also included for comparison. All configurations employ SRMs at both the local agent (i.e., DU) and centralized critic levels, with prompt generation and fusion extending the methodology established in prior work~\cite{lotfi2025llm}.

\begin{figure}[t!]
  \centering
\includegraphics[width=0.65\columnwidth]{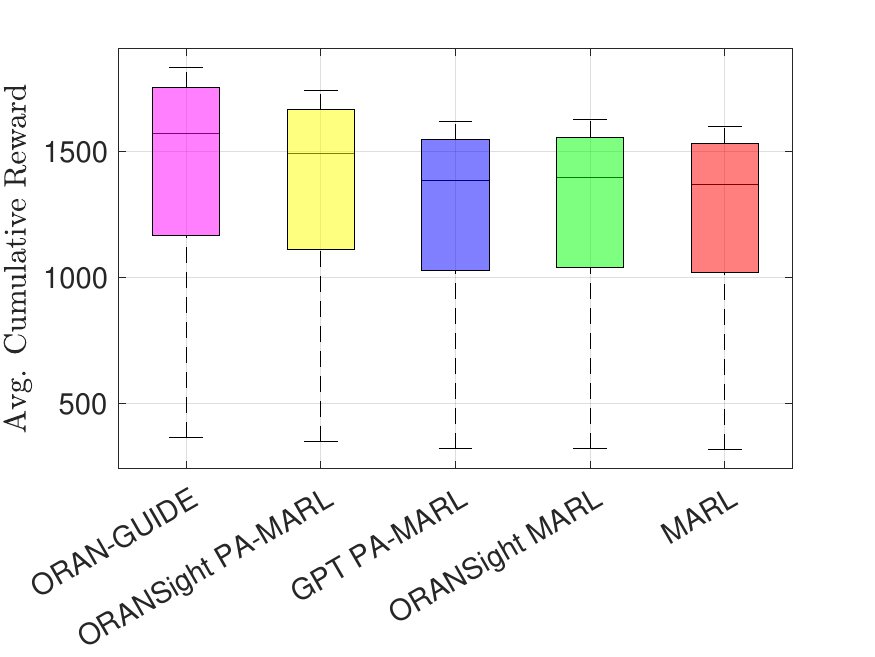}\vspace{-0.2cm}
    \caption{\small Statistical distribution of cumulative returns for each method.
    }\vspace{-0.2cm}
    \label{boxplot}
\end{figure}
\begin{figure}[t!]
  \centering
\includegraphics[width=0.65\columnwidth]{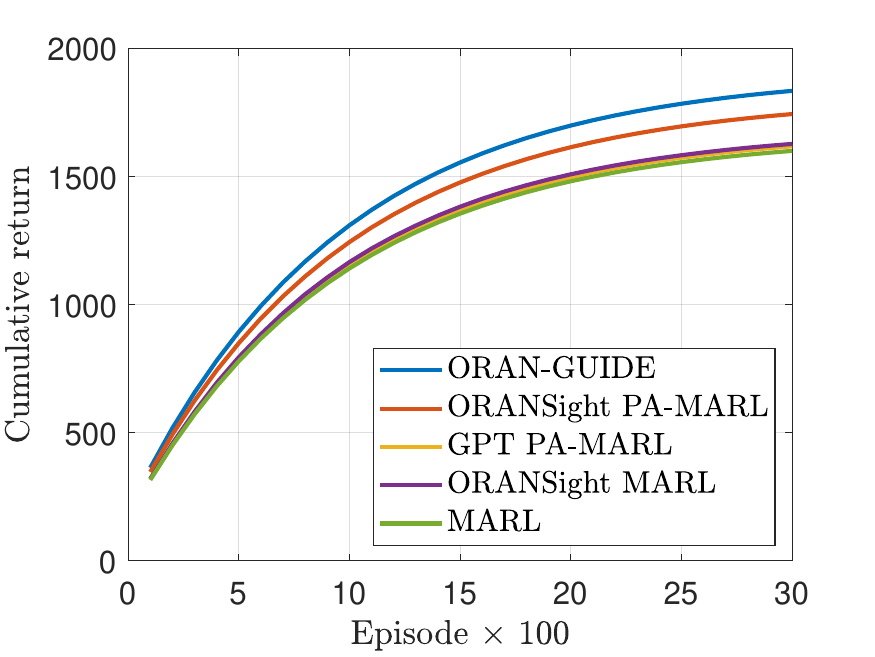}\vspace{-0.2cm}
    \caption{\small Comparison of average cumulative rewards across methods.  
    }\vspace{-0.2cm}
    \label{cum_rew}
\end{figure}
\subsection{Convergence Performance}
Fig.~\ref{cum_rew} illustrates the convergence performance of the proposed ORAN-GUIDE framework compared to other baselines. The results show that ORAN-GUIDE consistently achieves faster convergence and higher average cumulative rewards across training steps, demonstrating its improved policy learning capability. 
To further highlight these performance differences, Fig.~\ref{boxplot} presents the distribution of average cumulative rewards across multiple runs. The results show that ORAN-GUIDE not only yields the highest median performance but also exhibits lower variability compared to other methods. This underscores its robustness and sample efficiency in dynamic O-RAN environments and confirm that combining domain-aware and task-aware prompts within a dual-LLM architecture significantly enhances DRL performance.

\subsection{Sample Efficiency via Relative Performance Improvement}

To evaluate the sample efficiency of the proposed ORAN-GUIDE framework, we report the Relative Performance Improvement (RPI) metric, defined as:
\begin{equation}
\text{RPI} = \frac{\text{Acc-Reward}_m - \text{Acc-Reward}_b}{\text{Acc-Reward}_b},
\end{equation}
following~\cite{wang2024llm}, where $\text{Acc-Reward}_m$ and $\text{Acc-Reward}_b$ denote the accumulated rewards of the evaluated method $m$ and the MARL baseline $b$, respectively, over 1000 training steps. 
As summarized in the last column of Table~\ref{table:ablation}, ORAN-GUIDE achieves the highest RPI, indicating superior sample efficiency in the initial training phase. Methods that rely solely on domain knowledge (ORANSight MARL) or prompt tuning (GPT PA-MARL) yield modest improvements, while combining both, as in ORAN-GUIDE, significantly boosts convergence. These results highlight the advantage of dual prompting for rapid policy adaptation in complex wireless environments.

\subsection{Number of Learnable tokens}
\begin{figure}[t!]
  \centering
\includegraphics[width=0.65\columnwidth]{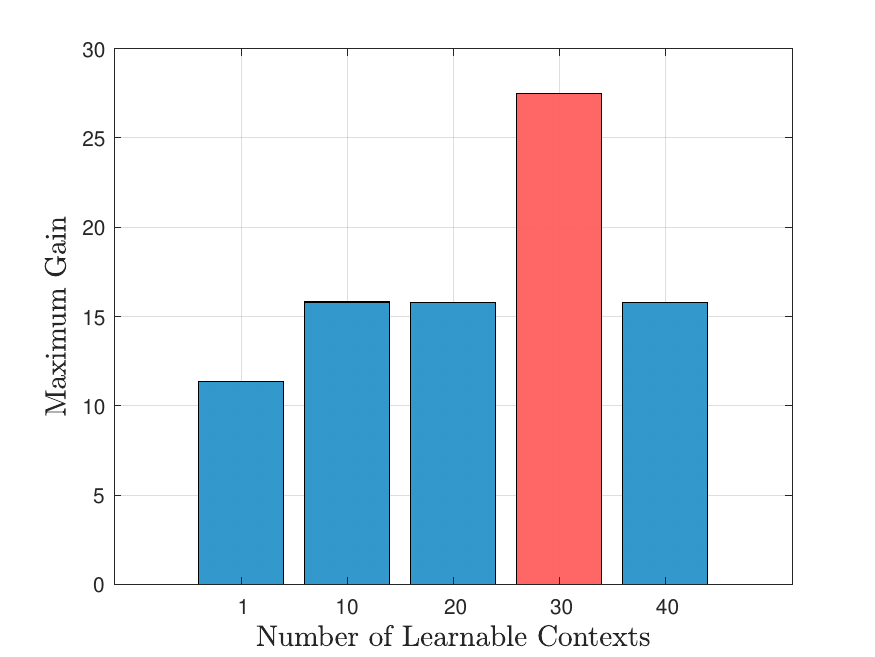}\vspace{-0.2cm}
    \caption{\small Comparison of maximum reward gains under varying numbers of learnable context.
    }\vspace{-0.2cm}
    \label{cntnumb}
\end{figure}

Figure~\ref{cntnumb} presents the performance impact of varying the number of learnable tokens (i.e., additional prompt tokens). Using too many tokens can lead to overfitting, whereas too few may limit the model’s capacity to capture task relevant semantics results in underfitting. As such, the token count should be treated as a tunable hyperparameter tailored to the task and deployment setting. The configuration yielding the highest performance is marked in red, indicating the optimal token range for this scenario. 

\begin{table}[t!] 
\footnotesize
\centering
\caption{Ablation Study Results }\vspace{-0.2cm}
\resizebox{\columnwidth}{!}{%
\begin{tabular}{|>{\centering\arraybackslash}m{2.4cm}|>{\centering\arraybackslash}m{1.2cm}|>{\centering\arraybackslash}m{1.2cm}|>
{\centering\arraybackslash}m{1.2cm}|>{\centering\arraybackslash}m{1.cm}|}
\hline
\textbf{Model Version}              & \textbf{QoS (eMBB)} & \textbf{QoS (mMTC)} & \textbf{QoS (URLLC)} & \textbf{RPI} \\ \hline
\textbf{ORAN-GUIDE}  & 87.7\%                & 62.0\%                 & 66.34\%                   & 14.74\%                \\ \hline
\textbf{ORANSight PA-MARL}  & 85.3\%                & 46.84\%                 & 25.42\%                   & 9.0\%                \\ \hline
\textbf{GPT PA-MARL}                & 34.4\%                & 21.57\%                 & 15.01\%                   & 0.84\%              \\ \hline
\textbf{ORANSight MARL}                & 24.1\%                & 2.63\%                 & 6.2\%                   & 2.16\%              \\ \hline
\end{tabular}%
}\label{table:ablation} \vspace{-0.3cm}
\end{table}

\subsection{Ablation Study}
Table~\ref{table:ablation} presents the results of an ablation study that isolates the contribution of each component and compares the proposed ORAN-GUIDE framework (cross-model reasoning), with its reduced variants: ORANSight PA-MARL (domain-specific reasoning only), which employ only one domain-specific LLM model, GPT PA-MARL (general reasoning only), which replaces the domain-specific LLM with a general-purpose GPT-2 model in the SRM; and ORANSight MARL, which removes the prompt-aligned LLM integration entirely. These are evaluated against a plain MARL baseline. These comparisons offer insights into how domain-aware prompt generation improves learning outcomes. 


\begin{figure}[h]
  \centering
    \includegraphics[width=0.78\columnwidth]{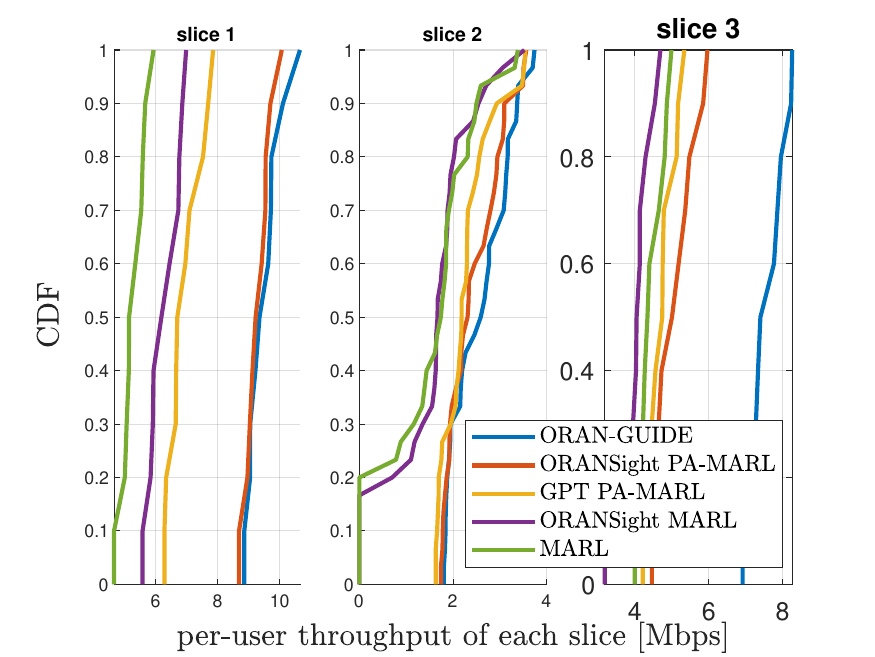}\vspace{-0.2cm}
    \caption{\small CDF of UEs throughput across the network.}\vspace{-0.2cm}
    \label{ue_throughput}
\end{figure}
\subsection{Network users' QoE}
Fig.~\ref{ue_throughput} illustrates the per-user throughput distribution (CDF) across the three slices. In all cases, the proposed ORAN-GUIDE framework achieves the highest throughput, demonstrating improved efficiency and fairness among users. ORANSight PA-MARL consistently performs second-best, benefiting from domain-specific prompting. GPT PA-MARL and ORANSight MARL offer moderate improvements over the baseline MARL, but show reduced consistency in comparison. These results underscore the importance of integrating domain and task-aligned prompt-driven LLM representations in achieving scalable and user-level performance gains in O-RAN slicing.

\section{Conclusions}\label{sec:conclusion}
Optimizing resource allocation and slicing in O-RAN environments requires robust frameworks that adapt to rapidly changing conditions, partial observability, and multi-slice service demands. To address these challenges, we introduced ORAN-GUIDE, a dual-LLM framework that enhances multi-agent reinforcement learning by generating semantically rich, task-aware state representations. By decoupling domain knowledge generation from policy learning, our approach enables distributed agents to make informed decisions with improved context awareness and generalization capabilities. Through the fusion of ORANSight-generated domain prompts and learnable task-adaptive tokens, the RL agents gain a deeper understanding of the network environment, allowing for more efficient and resilient policy optimization. Simulation results demonstrate that ORAN-GUIDE significantly outperforms conventional MARL and single-LLM baselines in terms of convergence speed, sample efficiency, and per-slice QoS. These findings highlight the promise of integrating LLMs with RL to build scalable, intelligent control systems for next-generation wireless networks.





\bibliography{Main}
\bibliographystyle{IEEEtran}



% ...

\end{document}